\documentclass[runningheads]{llncs}
\usepackage{graphicx}
\usepackage{booktabs}
\usepackage{amsmath,amssymb,amsfonts}
\usepackage{algorithmic}
\usepackage{textcomp}
\usepackage{xcolor}
\usepackage{booktabs}
\usepackage{lipsum,lineno}

\usepackage{todonotes}
\usepackage{svg}
\usepackage{array}\newcolumntype{P}[1]{>{\centering\arraybackslash}p{#1}}
\usepackage{ifthen}
\usepackage{xcolor}
\usepackage{cite}
\usepackage{verbatim}
\usepackage{placeins}
\usepackage{lscape}
\usepackage{multirow}
\usepackage{float}
\usepackage{lipsum}  
\usepackage{todonotes}
\usepackage{nicematrix}
\usepackage{hyperref}
\usepackage{enumitem}
\usepackage{siunitx}
\usepackage{adjustbox}
\usepackage{graphicx}

\definecolor{lavenderblush}{rgb}{0.93, 0.88, 0.90}%{1.0, 0.94, 0.96}

\colorlet{shadecolor}{lavenderblush}

\newcommand{\mytitle}{Reflective Dialogue or Prompt Refinement?
Effects of Tutor Scaffolding on Students’ Independent LLM Use for Programming}
\newcommand{\myshorttitle}{Reflective Dialogue or Prompt Refinement?}

\newboolean{ISanonymous}
\setboolean{ISanonymous}{False}
\ifthenelse{\boolean{ISanonymous}}{\newcommand*{\Anonymous}{}}

\newcommand{\KWA}{Large Language Models}
\newcommand{\KWB}{Computing Education}
\newcommand{\KWC}{Behaviors}
\newcommand{\KWD}{Performance}
\newcommand{\KWE}{Learning}
\newcommand{\KWF}{Effectiveness}
\newcommand{\KWG}{Tutor}

\newcommand{\MOBOTS}{MOBOTS \& LEARN, EPFL, Switzerland}

\newcommand{\HEPVaudEnglish}{University of Teacher Education (Haute École Pédagogique) Vaud, Switzerland}

\newcommand{\CEDE}{Center for Digital Education \& LEARN, EPFL, Switzerland}

\newcommand{\AuthorLH}{Laila El-Hamamsy}
\newcommand{\AuthorKU}{Kim Uittenhove}

\newcommand{\AuthorFM}{Francesco Mondada}
\newcommand{\AuthorAP}{Aitor Perez}
\newcommand{\AuthorPJ}{Patrick Jermann}

\newcommand{\AuthorEB}{Engin Bumbacher}

\newcommand{\AuthorJB}{Jérôme Brender}

\newcommand{\EmailLH}{laila.elhamamsy@hepl.ch}
\newcommand{\EmailPJ}{patrick.jermann@epfl.ch}
\newcommand{\EmailAP}{aitor.perez@epfl.ch}
\newcommand{\EmailKU}{kim.uittenhove@epfl.ch}
\newcommand{\EmailEB}{engin.bumbacher@hepl.ch}

\newcommand{\EmailJB}{jerome.brender@epfl.ch}

\newcommand{\EmailFM}{francesco.mondada@epfl.ch}

\newcommand{\ORCIDLH}{0000-0002-6046-4822}
\newcommand{\ORCIDKU}{0000-0001-5450-3875}

\newcommand{\ORCIDEB}{0000-0002-4322-7059}

\newcommand{\ORCIDPJ}{0000-0001-9199-2831}
\newcommand{\ORCIDAP}{0000-0001-6412-3235}

\newcommand{\ORCIDFM}{0000-0001-8641-8704}

\newcommand{\ORCIDJB}{0009-0008-3279-5641}

\begin{document}

\title{\mytitle\thanks{We thank our colleagues (M.B, A.B, L.C, E.C, M.D, A.S, G.D, V.D, F.E) for their support, and the students who generously volunteered to participate in the study.}
}
\titlerunning{\myshorttitle}

\ifdefined\Anonymous 

    \author{Anonymous 1\inst{1,2} \and
    Anonymous 2\inst{2} \and
    Anonymous 3\inst{1} \and
    Anonymous 4\inst{1} \and
    Anonymous 5\inst{2} \and
    Anonymous 6\inst{1} \and
    Anonymous 7\inst{1}
    }
    \authorrunning{Anonymous 1 et al.}
    \institute{Affiliation 1
    Affiliation 2
    }

\else

    \author{\AuthorJB\inst{1,2}\orcidID{\ORCIDJB} \and
    \AuthorLH\inst{2}\orcidID{\ORCIDLH} \and
    \AuthorKU\inst{1}\orcidID{\ORCIDKU} \and
    \AuthorAP\inst{3}\orcidID{\ORCIDAP} \and
    \AuthorPJ\inst{3}\orcidID{\ORCIDPJ} \and
    \AuthorFM\inst{1}\orcidID{\ORCIDFM} \and
    \AuthorEB\inst{2}\orcidID{\ORCIDEB} 
    }
    \authorrunning{J. Brender et al.}
    \institute{
    \MOBOTS \\ \email{\EmailJB, \EmailKU,  \EmailFM} \and
    \HEPVaudEnglish \\ \email{\EmailEB, \EmailLH} \and
    \CEDE \\ \email{\EmailAP, \EmailPJ}
    }
    
\fi

\maketitle

\begin{abstract}
    %%%%%%%% MAX 215 words OTHERWISE IT GOES OVER THE PAGES %%%%%%%%
While Large Language Models (LLMs) can provide personalized support in learning, several studies have raised concerns regarding their use in education. Importantly, learning depends on how students engage with LLMs. This study examined how two types of LLM-based tutors shape students’ prompting practices, learning, and subsequent LLM-use: a Socratic-Guidance (SG) tutor, which structures interaction through dialogic questioning, and a Prompt-Refinement (PR) tutor that guides the formulation of effective prompts. We conducted a two-phase study in a graduate-level mobile robotics course: 66 students used either the SG or PR tutor during a 6-week intervention, followed by 52 students using an unconstrained LLM during a 3-week course project. 
Results show that while the SG- and PR tutors led to similar task performance and prompting patterns during guided use, they differ in learning outcomes and later LLM-use. SG-students, relative to PR-student, achieved higher learning gains in later sessions, and were more likely to adopt understanding-driven prompting strategies, which are predictive of higher understanding, when using an unconstrained LLM. Although learners perceived the SG tutor as less efficient, the findings suggest that Socratic guidance supports the development of students’ capacity to learn with LLMs over time, highlighting its importance for LLM tutor design.
%%%%%%%% MAX 215 words OTHERWISE IT GOES OVER THE PAGES %%%%%%%%

\begin{comment}
Prior work suggests that constrained and Socratic tutoring approaches can support learning during interaction by shaping the AI’s responses. However, their impact on students’ prompting practices remains unclear, as does whether such approaches foster durable and transferable ways of interacting with LLMs once constraints are removed. This gap highlights a design distinction between systems that guide learning through the tutor’s behavior and systems that explicitly scaffold how students formulate prompts.

We address this gap through a mixed-methods experimental study examining two design approaches: (i) a Socratic-Guidance (SG) tutor and a Prompt-Refinement (PR) tutor, and their effects on prompting behaviors, task performance, and learning; and (ii) the persistence of prompting behaviors when students subsequently interact with an unconstrained LLM, and their relation to learning outcomes.

During six weeks, 66 students participated in practice lab sessions and were randomly assigned to either the SG condition (n=29) or the PR condition (n=37). During a subsequent three-week project phase, 52 students used with an unconstrained GPT system. The two Ai tutor designs contrasted (i) modifying how the AI responds in order to support reflective dialogue (SG) with (ii) supporting how students formulate their prompts (PR). Data sources included pre- and post-tests of learning, scores practice labs, project grades, prompt interactions, and perception questionnaires.
\end{comment}
    \keywords{\KWA \and \KWB \and \KWC \and \KWD \and \KWE \and \KWF \and \KWG}
\end{abstract}

%How humans collaborate with generative AI, particularly in terms of productivity. Findings suggest that access to these tools allows people/workers/ to perform better in productivity and tasks more efficiently

\vspace{-10pt}
\section{Introduction}
\vspace{-5pt}

With the rise of Large Language Models (LLMs), a growing body of work has examined applications of LLM-based tools in education, including curriculum development, automated assessment and feedback, and tools that provide personalized support during learning activities \cite{kasneci_chatgpt_2023,shahzad_comprehensive_2025}. At the same time, researchers have argued that educational benefits are not inherent to such LLM-based systems but depend on their pedagogical design and integration, and have warned of substantial risks when these conditions are not met \cite{kasneci_chatgpt_2023}.

\vspace{-5pt}\subsection{Cognitive and metacognitive risks of LLM use in learning}\vspace{-5pt}

One central risk associated with students’ use of LLM-based systems is the conflation of improved task performance with learning. While studies have found that students who use such systems to solve problems often achieve higher performance, these gains are neither indicative of students’ ability to perform without the tool\footnote{Studies have found that when access to LLMs is removed, students' performance declines significantly, and even ``perform worse than those who never had access'' \cite{bastani2024generative}} nor of improved learning outcomes\footnote{Studies have found weak or no correlation between performance during LLM-use and subsequent learning \cite{brender2024s, fan2025beware}}. 

This disconnect between performance and learning can be partly explained through research on cognition and metacognition. Several studies have raised concerns that students’ use of LLMs may foster metacognitive laziness \cite{sabqat_shortcut_2025, delikoura_superficial_2025, fan2025beware}. As LLMs become more capable, students may increasingly rely on these systems in ways that reduce their active cognitive and metacognitive engagement \cite{zhai2024effects}. Evidence for such effects is provided by a recent EEG study in the context of a writing task \cite{kosmyna2025your}: LLM-supported participants achieved higher task performance, but the authors report a resulting ``cognitive debt'', reflected in reduced critical evaluation of LLM outputs and impaired long-term semantic retention.

Importantly, these risks are not inherent to LLM-based systems but depend on how interaction with them is structured in learning contexts. Prior work shows that students adopt diverse LLM-use patterns, which are associated with different performance and learning outcomes \cite{brender2024s}. For example, learning benefits are more likely when students engage LLMs through conceptual questioning \cite{brender2024s} or use prompts that foreground understanding rather than solution generation, such as prompts focused on reasoning about code \cite{brender2025structured}.

\vspace{-10pt}\subsection{Improving the effectiveness of LLMs for learning}\vspace{-3pt}

Thus, if the goal is to support productive LLM use beyond instructional settings, then the central design challenge is how to help students develop productive ways of engaging with LLMs that persist beyond instruction. Prior work has pursued two broad approaches toward this end.

\vspace{-10pt}\subsubsection{Socratic scaffolding for reflective LLM use.}

One approach to supporting student learning with AI-based systems has been to constrain assistance to explanations \cite{okonkwo2020python}, hints \cite{yang2024enhancing}, or course-aligned guidance \cite{okonkwo2020python, liu2024teaching} rather than providing direct answers, thus encouraging students to actively construct solutions. A prominent version of this design principle is Socratic questioning \cite{chen_exploring_2025}, which relies on question-driven dialogue to promote reflective cognitive processes \cite{paul2007critical}.

When implemented in LLM-based tutors, Socratic scaffolding has been shown to support deeper reasoning and understanding \cite{favero2024enhancing, shetye2024evaluation, goyal2025sakshm}, though effects are not uniform across students, which highlights the need to combine this approach with other instructional strategies \cite{kampylis2026timeless}. Related work further underscores the importance of design choices: One study found that access to an unguarded generative AI interface can improve short-term performance, but it may lead to worse subsequent performance once access is removed, an effect that in turn is mitigated when pedagogical safeguards are in place \cite{bastani2024generative}.

These findings highlight the need to examine whether system-side scaffolding strategies, such as Socratic questioning, also support students in developing productive LLM use beyond structured instruction.

\vspace{-7pt}\subsubsection{Teaching prompting for productive LLM use.}

In contrast to Socratic scaffolding, which structures how the LLM responds to students, a second approach seeks to shape how students engage with LLMs. This line of work focuses on teaching students how to formulate prompts and interact with LLMs in ways that support productive use beyond instructional settings \cite{mollick_assigning_2023}.

One strand of work has explored structuring chatbot interfaces to guide prompt formulation \cite{brender2025structured}. While this approach has been found to improve prompting behaviors in the moment, these effects did not persist after removing constraints. Also, students resisted restrictive interfaces, despite recognizing potential benefits, and may actively bypass constraints to obtain full responses \cite{brender2025structured, kapoor2025exploring}.

A second strand focuses on teaching prompting strategies through guidelines or instruction. LLM output quality depends strongly on users’ inputs \cite{anthropic2026aeiv4}, yet effective prompting remains challenging, particularly for learners with limited domain expertise \cite{zamfirescu2023johnny}. In response, numerous prompting guidelines have been proposed \cite{denny2023promptly, mollick2022new, stokel2023chatgpt}; however, many lack empirical validation and rarely examine whether learned strategies transfer to unconstrained LLM use \cite{yang2023use, brender2025structured}.

Research needs to examine whether such learner-side strategies support productive LLM use without fostering over-reliance or superficial adoption.

\vspace{-5pt}\subsection{Present Work}

These strands of work raise complementary questions about whether productive LLM use is better supported by structuring how systems respond to students or by teaching students how to engage with LLMs themselves. The present study addresses this question by directly comparing two contrasting design approaches for supporting students’ interaction with LLMs: (1) a Prompt-Refinement (PR) tutor that supports students in formulating prompts conducive to learning, and (2) a Socratic-Guidance (SG) tutor, which represents a state-of-the-art system  that structures the system responses to support student reflection. Building on the literature reviewed above, we address the following research questions (RQs):
\vspace{-8pt}
\begin{itemize}
    \item[(RQ1)] How do the SG and PR tutors shape students’ prompting behaviors, task performance, and learning during guided lab sessions?
    \item[(RQ2)] How do prompting practices relate to task performance and learning?
    \item[(RQ3)] How do students engage with an unconstrained LLM after the intervention?
\end{itemize}

We investigate these research questions in a mixed-methods study conducted in a graduate-level mobile robotics course. During a six-week intervention phase, students were randomly assigned to either the SG- or PR-tutor condition across all practical lab sessions. This was followed by a three-week project phase in which all students interacted freely with an unconstrained LLM.

\vspace{-5pt}
\section{System Design and AI-tutor Variants}
\label{sec:system_design}
\vspace{-5pt}

The study employed a web-based chatbot integrated with OpenAI’s API (GPT-5) to provide real-time assistance to students. The system is built on a Retrieval-Augmented Generation (RAG) architecture using all course materials, including lecture slides, forum Q\&A archives, exercises, and solutions. Rather than generating responses in a single step, each student prompt is processed through a multi-step, agentic workflow that enables control over retrieval, response generation, and the timing of pedagogical constraints. Using this architecture, we implemented two AI tutor variants that differ in how interaction is structured.

\vspace{-10pt}\subsubsection{The Prompt-Refinement (PR) tutor.} This tutor was designed to support students in crafting high-quality prompts by requiring prompt refinement before any assistance is provided \cite{lo2023clear}. For each incoming prompt, the system evaluates prompt quality along two dimensions identified in prior works \cite{ekin2023prompt,brender2025structured}: (1) \emph{Clarity and specificity}, i.e., whether the prompt provides sufficient context, constraints, and expected outputs; and (2) \emph{Learning intent and reasoning}, i.e., whether it articulates the student’s understanding or poses a focused conceptual question rather than requesting a solution. If a prompt is insufficient on either dimension, the PR tutor interrupts the interaction and provides brief rubric-based feedback along with two alternative prompt suggestions. Students may select one of these or revise their original prompt. Only once the revised prompt meets the minimum criteria does the system generate a response.

\vspace{-10pt}\subsubsection{The Socratic-Guidance (SG) tutor.} This tutor was designed to scaffold learning by shaping how the system responds to student queries. Rather than providing direct solutions, the tutor generates open-ended, reflective questions that prompt students to clarify their understanding and advance their reasoning toward a solution \cite{kestin2025ai, vanzo2025gpt, favero2024enhancing}. This questioning stance is maintained throughout the interaction, in line with established guidelines for educational AI use \cite{OpenAI_teaching_2024}.

\begin{comment}
    
\vspace{-5pt}\subsection{Tutor refinement for in class use}
%if no space to remove 
The system prompts were refined for both types of tutors using two years of historical interaction data and pre-validated with six teaching assistants who tested the tutors prior to the intervention.
For both conditions, a two-phase scaffolding strategy was employed:
Phase 1 (Weeks 2–4) enforced strict constrained without direct solutions;
Phase 2 (Weeks 5–7) allowed direct answers for basic syntax and short code snippets while for the SG-tutor preserving guided questioning for higher-level reasoning. %The system prompt is provided at the following link: [Anon. ins]
\end{comment}

\vspace{-10pt}\section{Methodology}

\vspace{-5pt}\subsection{Study Context and Design}
\vspace{-5pt}

\begin{figure}
    \centering
    \vspace{-5pt}
    \includegraphics[width=\linewidth]{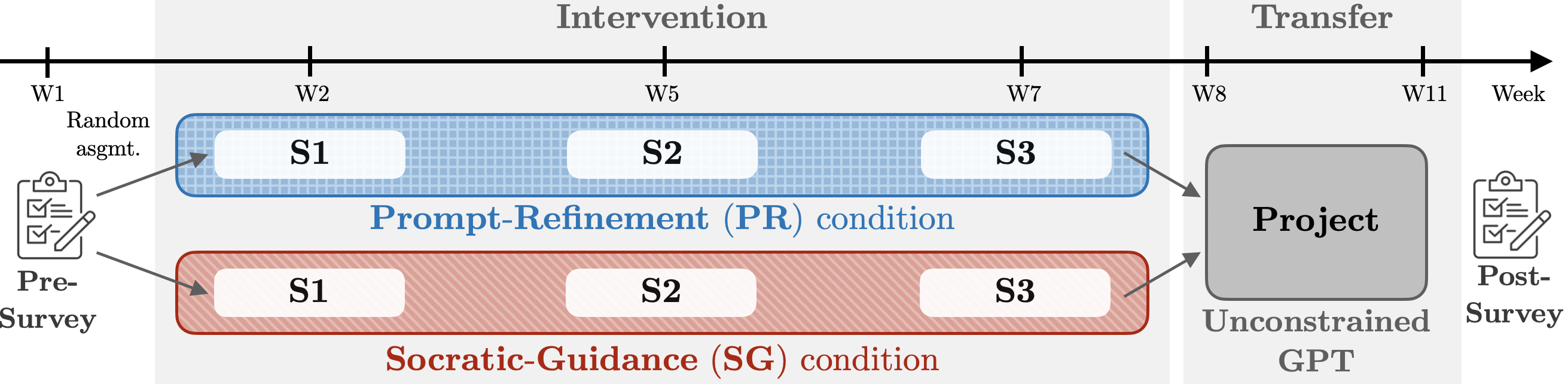}
    \vspace{-15pt}
    \caption{Overview of the study design. %Participants were randomly assigned to the Socratic Guidance (SG) or Prompt Reformulation (PR) conditions for the 6-week intervention phase. 
    %A pre-survey was administered at week 1, followed by a post-survey at the conclusion of the 11-week period. Data collection throughout 3 of the 6 sessions included learning assessments (pre/post-tests), prompt logs, and task performance metrics.
    }
    \label{fig_study_design}
    \vspace{-15pt}
\end{figure}

The study (see Fig.~\ref{fig_study_design}) was conducted in a graduate-level mobile robotics course at \ifdefined\Anonymous Anonymous University \else EPFL\fi \footnote{The study was approved by \ifdefined\Anonymous Anon. Ethical Committee\else EPFL's Ethics Committee\fi \ifdefined\Anonymous\else (HREC000658/15.07.2025)\fi} and followed a mixed-methods, between-subjects design. %Participants were randomly assigned to either a Socratic-Guidance (SG) or Prompt-Refinement (PR) tutor condition.

\vspace{-5pt}\subsection{Study Procedure}
\vspace{-5pt}

%Prior to the intervention, students completed a perception survey (see section~\ref{sec:data_collection})%, and were then randomly assigned to one of the two conditions. %The intervention took place over the course of three lab sessions (S1–S3). Each lab session started and ended with a content-related pre-test and post-test. Following the lab phase, students completed a team-based project during which they interacted freely with an unconstrained LLM.

%and a pre-test on the content  and were randomly assigned to either the SG ($n=29$) or PR ($n_{lab}=37$, $n_{project}=23$) condition.
%Prior to the intervention, students completed a perception survey (see Section~\ref{sec:data_collection}). The intervention took place during the lab component of the course and included three designated measurement sessions (S1–S3). Following the lab phase, students completed a team-based project during which they interacted freely with an unconstrained LLM.
As shown in Figure \ref{fig_study_design}, students started with a perception survey (see Section~\ref{sec:data_collection}) and were then randomly assigned to one of the two AI tutor conditions. The intervention consisted of three lab sessions (S1–S3), during which students interacted with the assigned tutor while working on programming tasks.

Each lab session followed the same structure: a content-related pre-test, a 75-minute hands-on programming phase during which students could use the chatbot for assistance, and a content-related post-test. Prompt logs and submitted code were collected during the programming phase.

Following the three lab sessions, students worked in teams of four on a course project over a three-week period. During this project phase, students could optionally interact with an unconstrained version of the course chatbot, using the same RAG system architecture but without the tutoring scaffolds described in Section~\ref{sec:system_design}. At the end of the course, students completed a post-survey assessing their perceptions and use of LLMs.

% Three out of the 6 lab sessions, each two weeks apart, were designated as measurement sessions (referred to S1, S2, and S3) as the task complexity made the LLM support pedagogically relevant (see section \ref{sec:data_collection}). 
% Each lab session was structured as follows: 
% (i) a 15-minute briefing, including a 10-minute pre-test;
% (ii) 75-minutes of hands-on programming tasks, where we collected prompt logs, and implemented codes, %with each task graded on a 0, 0.5, or 1 scale; and%. The complexity of the tasks made the use of ChatGPT pedagogically meaningful, and
% (iii) a 10-minute post-test%with in-depth questions covering the same content
% .

%Post-intervention, the students worked in groups of 4 on a three week course project. During this project phase, all participants interacted freely with an unconstrained LLM, using the same system and interface as during the intervention but without pedagogical constraints. A post-survey was administered at the end of the course to assess changes in students’ perceptions and use of LLMs.

\vspace{-5pt}\subsection{Participants}
\vspace{-5pt}

Participants were recruited from the course cohort. This study focuses on session 3 and the course project. Of the 172 enrolled students, 66 participated in the lab session 3 ($n_{male}=47$, $n_{female}=15$, $n_{undisclosed}=4$). Then, 52 students volunteered to use the unconstrained course chatbot (RAG without scaffolds) during the course project ($n_{male}=33$, $n_{female}=15$, $n_{undisclosed}=4$). Participation was voluntary, based on informed consent, and financially compensated.

\vspace{-7pt}\subsection{Data Collection and Coding}
\label{sec:data_collection}

\vspace{-5pt}\subsubsection{The pre- and post-surveys (perception)} comprised multiple instruments. This paper focuses on analyses of programming experience, perceptions of LLMs \cite{teo2009modelling}, and self-reported LLM usage.\footnote{The complete survey is accessible \href{https://doi.org/10.5281/zenodo.19260026}{here}.}%thatencompasses five dimensions: utility, ease of use, attitude/interest, self-efficacy, and intent to employ in future teaching activities\footnote{The complete survey is accessible \href{https://drive.switch.ch...}{here}.}. 

\vspace{-15pt}\subsubsection{The lab session pre-tests (learning)} \label{Pre_Test}
evaluated two components: (i) algorithmic explanation (3 pts) and (ii) procedural sequencing of the algorithm targeting key theoretical concepts from the lab (3 pts). Open-ended responses were anonymized and graded by two Teaching Assistants (TAs). Inter-rater reliability (IRR) reached substantial levels across the full dataset (Fleiss' $\kappa_{S1}=0.79$, $\kappa_{S2}=0.70$, $\kappa_{S3}=0.77$).
We verified that students in both conditions had similar prior knowledge (S1: $U=327.5, p=.18, M=44.7\pm 18.0\%$; S2: $U=369, p=.54, M=61.5\pm 21.3\%$; S3: $U=380, p=0.06, M=54.5\pm 24.5\%$), we observed a marginal trend toward a lower score for S3, which is further discussed in the result section.

\vspace{-15pt}\subsubsection{The lab session post-tests (learning)}
consisted of multiple-choice questions targeting the same concepts as the pre-test (S1: 5 items; S2: 8 items; S3: 5 items). In addition, students completed a procedural sequencing task similar to the pre-test. %, targetting key theoretical concepts from the lab. 
Each MCQ was one point, with an additional 25\% weighting applied to the sequencing task. Scores were summed and standardized to compute an overall post-test score per session ($M_{S1}=44.4\pm 18.6\%$, $M_{S2}=45.5\pm 32.2\%$, $M_{S3}=64.9\pm 16.8\%$).

\vspace{-15pt}\subsubsection{The project understanding scores (learning)} were assigned to each group of 4 students by two experts based on a 45-minute oral exam assessing the students’ understanding of the underlying course theory used in the project.

\vspace{-15pt}\subsubsection{The practice lab scores (task performance)} were evaluated based on the practice lab tasks. Each task %(9 in S1, 5 in S2, and 8 in S3) 
was graded as 0 for incomplete, 0.5 for over 70\% correct, and 1 for fully correct. 
To assess IRR, 20\% of the submissions were %randomly sampled and 
graded by three TAs, yielding substantial agreement (Fleiss’ $\kappa_{S1} \in [0.72, 0.81]$; $\kappa_{S2} \in [0.67, 0.78]$; $\kappa_{S3} \in [0.70, 0.78]$). The remaining submissions were then distributed and graded individually. The total performance score was computed as the standardized sum of individual task scores.

\vspace{-15pt} \subsubsection{The chatbot interactions and prompt coding (prompting)}
comprised all student--chatbot interactions collected during the final lab session (S3) and the project phase, totaling 599 student prompts from S3 and 1120 student prompts from the project
($M_{S3}=9.1 \pm 5.8$; $M_{project}=21.5 \pm 27.1$ prompts per student). Prompts unrelated to task completion were excluded.

To characterize students' engagement with the chatbot systems, prompts were coded along two dimensions based on prior work \cite{brender2024s,brender2025structured}:

\begin{enumerate}[leftmargin=1.5em, noitemsep, topsep=3pt, label=\arabic*)]
  \item \textit{Prompt type} (role of the prompt): comprises \textit{implementation} (code generation or complete solutions), \textit{debugging} (error identification or resolution), and \textit{conceptual} (code comprehension or underlying computing concepts).
  \item \textit{Prompt quality} (how requests were formulated): \textit{understanding} (explicit explanations requests), \textit{granularity} (use of specific contextual details like variables or specific code), and \textit{clarity} (well-specified/unambiguous instructions).
\end{enumerate}

Three Teaching Assistants (TAs) independently annotated 20\% of the prompts, achieving substantial agreement
(Fleiss' $\kappa_{\text{prompt-type}}=.72$; $\kappa_{\text{prompt-quality}}\in [0.62, 0.80]$),
before coding the remaining data individually.
Prompt-level codes were aggregated at the student and group levels for subsequent analyses.

%Two additional categories were introduced especially in the context of the project-level reasoning. \textit{System Integration} refers to prompts that coordinate multiple modules within the prompt (e.g., localization + control, computer vision + path planning). 
%\textit{Planning} refers to prompts aimed at identifying and structuring the steps needed to solve a problem, prior to concrete implementation. Consistent with prior analyses of student help-seeking behavior \cite{phung2025plan}, this category captures prompts focused on design decision-making including selecting between alternatives, reasoning about trade-offs, or outlining implementation steps.
Finally, two TAs collaboratively annotated student-authored portions of each prompt to identify copy-pasted segments. These annotations were used to compute the proportion of personally written text per prompt.

\vspace{-10pt}\subsection{Data Analysis}
\label{sec:data_analysis}
For the scope of this paper, analyses of process data focus on data from the third lab session (S3) and the course project, which together capture students’ behavior after sustained exposure to the tutor and during subsequent unconstrained LLM use.

\vspace{-12pt}\subsubsection{Prompting patterns during and after intervention.}\label{clustering}

Given the multi-dimensional nature of the prompt type and quality metrics, we conducted separate cluster analyses on data from S3 and the project to identify recurring prompting patterns. Clustering captured students’ relative use of different prompt dimensions when using the chatbots. We iteratively evaluated different feature combinations and clustered students following the procedure described in \cite{shved2024teaching}.

First, for each feature, students were split into high- and low-use groups based on pairwise Euclidean distances between students, which were then converted into similarity matrices using a Gaussian kernel ($\sigma = 0.0001$).
Second, spectral clustering was applied to different feature combinations, with the optimal number of clusters selected using the Silhouette score.
Third, clusters were labeled based on the prevalence of high- versus low-use students for each feature.

Finally, the feature combination that maximized the Silhouette score was selected. In particular, aggregating \textit{conceptual} and \textit{understanding} prompts using a union-based (OR) aggregation into a single dimension for \textit{Understanding}, and aggregating \textit{clarity} and \textit{granularity} into a single dimension for prompt \textit{Quality}, consistently improved cluster cohesion and separation.
The final feature representation thus comprised four dimensions: three prompt-type dimensions (\textit{Implementing}, \textit{Debugging}, and \textit{Understanding}) and one prompt \textit{Quality} dimension.

This clustering procedure resulted in four prompting pattern types in S3 (average silhouette score $s=0.74$) and three in the project phase ($s=0.64$).

\vspace{-14pt}\subsubsection{Task performance and learning during intervention.}

To examine in-session task performance and learning outcomes, we used linear mixed-effects models for practice lab scores and post-test results.
Post-test scores were analyzed using a repeated-measures design, with students modeled as a random effect. Task performance was analyzed using mixed-effects models with pre-test scores as a covariate. All continuous variables were standardized, allowing regression coefficients ($\beta$) to be interpreted as effect sizes. Model assumptions were evaluated by inspecting residuals, with Shapiro-Wilk tests used to assess normality.

\vspace{-14pt}\subsubsection{Post-intervention project understanding score.} 
Project understanding was assessed at the group level for 30 project groups, as scores were assigned based on a group oral examination rather than individual performance. Groups varied in their composition of intervention students (SG or PR) and non-inter-vention students, and only a subset of intervention students used the chatbot, resulting in partial observability of prompting patterns at the group level.

To account for this partial trace coverage, we fitted a linear regression model predicting the non-standardized group-level understanding score from (a) the total number of intervention students, (b) the number of intervention students without chatbot use, and (c) the number of students assigned to each intervention condition or prompting-pattern cluster.

\vspace{-10pt}\section{Results}

\begin{comment}
\subsection{Controlling for inter-condition population differences}

Provided the lack of control group, it was important to ensure that the students in both conditions were similar before the start of the intervention. 

To that effect, we compared the distribution of student responses on the pre-survey data using Mann Whitney U tests. The results indicate that students in the Socratic and PR conditions did not differ ($p>0.05$) in terms of\todo{@Engin: do I run the test on the average of each metric and report that?}: 

\begin{itemize}
    \item Self-reported prior programming experience, self-efficacy using LLMs, perceived prompt usage (in terms of prompt types and quality). 
    
    \item XXX\todo{Add a descriptor of what these metrics intend to encompass} metrics pertaining to elaboration\todo{Need to be more explicit about what this is}, mastery motivation, and performance motivation. 
\end{itemize} 

We further validated that the populations did not differ prior to the interventions by comparing the pre-test data using Kruskal-Wallis tests ($H_{s1}=$, $p_{s1}=XXX$; $H_{s2}=$, $p_{s2}=XXX$; $H_{s3}=$, $p_{s3}=XXX$)\todo{Insert the values}. \todo{Issue here, it contradicts the regression model for s3 where we see that being in the socratic condition in pre-test is significantly lower...}
\end{comment}

\vspace{-6pt}\subsection{Effect of the intervention on prompt type and prompt quality}\vspace{-5pt}
\label{sec:intervention_prompt}

As described in Section~\ref{clustering}, the cluster analysis of S3 data across conditions identified four distinct prompting patterns. Based on their high--low distribution of prompt types and prompt quality, these were labeled as (i) \textit{Implementers} with low-quality prompts ($n=14$), (ii) \textit{Debuggers} with high-quality prompts ($n=11$), (iii) \textit{Understanding-driven} prompters with high-quality prompts ($n=8$), and (iv) \textit{Implementers} with high-quality prompts ($n=33$).

A $\chi^2$ test examining the distribution of students across prompting-pattern clusters by intervention condition revealed no significant differences between the SG- and PR-tutor conditions ($\chi^2=1.12$, $p=0.77$). This suggests that in the third lab session, the type and quality of prompting patterns enacted by students did not differ as a function of tutor condition.

\vspace{-8pt}\subsection{Effect of the intervention on task performance and learning}\vspace{-5pt}

We examined the impact of the SG- and PR-tutors on students’ task performance during the lab sessions and on learning outcomes measured by pre-post tests across sessions. To ensure comparability across sessions, all analyses were restricted to the 66 students who attended session S3.

\vspace{-8pt}
\paragraph{Task performance:} For each session, we fitted regression models with task performance as the dependent variable and condition and pre-test scores as predictors. Results showed no significant effect of condition on task performance in sessions S2 and S3 (no significant main or interaction effects).
In contrast, in session S1, students in the SG condition exhibited significantly lower task performance than those in the PR condition (main effect of SG condition: $\beta_{S1}=-0.8$, $p=0.001$).

This initial task performance decrement suggests that students required time to familiarize themselves with the SG-tutor, highlighting the importance of repeated exposure to support effective use of the system.
\vspace{-14pt}
\begin{figure}
    \centering
    \includegraphics[width=0.63\linewidth]{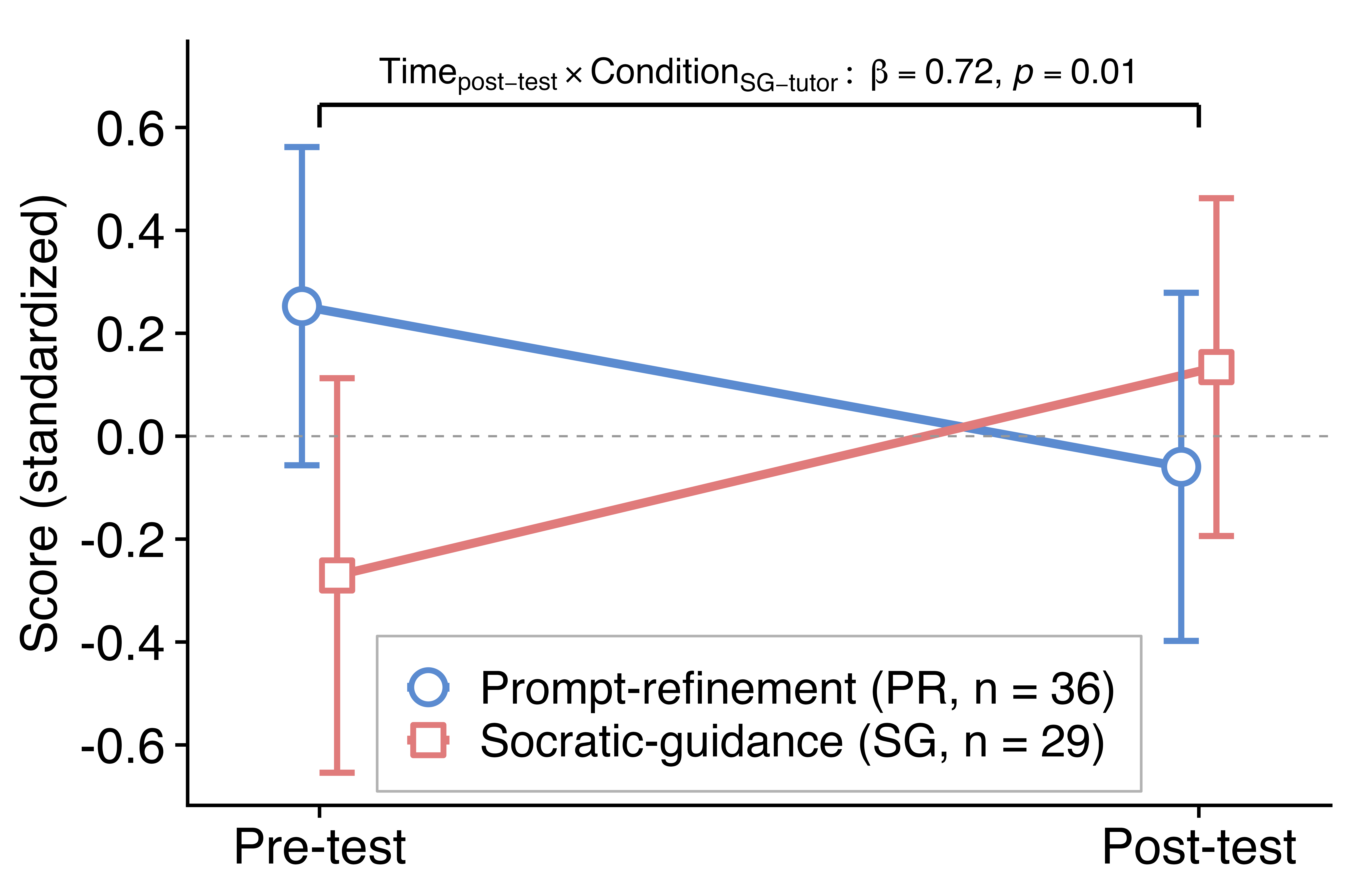}
        \caption{Learning outcomes for S3: Mean standardized scores by condition and time (pre--post). Error bars represent 95\% confidence intervals. Only the significant interaction from repeated-measures mixed-effects model is indicated.}
    \label{fig:learning_s3}
\end{figure}

\paragraph{Learning outcomes:} Learning outcomes were analyzed using repeated-measures mixed-effects models for each session, with test score as the dependent variable and the main effects as well as the interaction between condition and time (pre-post) as the predictors. Results indicated no significant effects (main and interaction) of condition in sessions S1 and S2.

In session S3, however, a significant interaction between time and condition was observed (Table~\ref{tab:learning_s3}).
While students in the SG condition had significantly lower pre-test scores (main effect of condition: $\beta=-0.52$, $p=0.02$), they showed significantly greater gains from pre- to post-test than students in the PR condition (interaction effect: $\beta=0.72$, $p=0.01$), see Figure~\ref{fig:learning_s3}.

\begin{table}
\vspace{-10pt}
\caption{Repeated-measures mixed-effects model of test scores by Time (pre--post), Condition (SG vs.\ PR), and their interaction. Estimates are standardized coefficients ($\beta$). Reference levels: Time = pre-test; Condition = PR-tutor.}
\label{tab:learning_s3}
\centering
\begin{tabular}{lccc}
\toprule
Predictors & Estimates & CI (95\%) & p \\
\midrule
(Intercept) & 0.25 & -0.06 -- 0.56 & 0.11 \\
Time [post-test] & -0.31 & -0.70 -- 0.07 & 0.11 \\ %C(time, Treatment(reference='pre_test'))[T.post_test] 
Condition [SG-tutor] & {-0.52}$^*$ & -0.98 -- -0.07 & 0.02 \\ %C(condition, Treatment(reference=0))[T.socratic]  
Time [post-test] : Condition [SG-tutor] & {0.72}$^*$ & 0.14 -- 1.30 & {0.01} \\ %C(time, Treatment(reference='pre_test'))[T.post_test]:C(condition, Treatment(reference=0))[T.socratic]
%Group Var & 0.XX &  ---  & --- \\
\midrule
\multicolumn{4}{l}{Observations (students): \hspace{0.0cm} 130 (65) \hspace{0.05cm} $R^2_\mathrm{marginal}$ / $R^2_\mathrm{conditional}$: 0.041 / 0.230 \hspace{0.0cm} \quad $^* p \le 0.05$} \\

\bottomrule
%\multicolumn{4}{c}{\quad $^* p \le 0.05$ \quad $^{**} p \le 0.01$ \quad $^{***} p \le 0.001$} \\
\end{tabular}
\vspace{-15pt}
\end{table}

In summary, these results indicate that while students using the SG-tutor initially experienced lower task performance, sustained use of the SG-tutor was associated with superior learning gains by session S3, without differences in task performance relative to the PR condition.

\vspace{-7pt}\subsection{Impact of prompt quality on task performance and learning}\vspace{-5pt}

To examine the relation of specific prompting patterns with task performance or learning, we analyzed the relationship between the prompting-pattern clusters in session~S3 (Section~\ref{sec:intervention_prompt}) and students’ task performance and learning outcomes.

\vspace{-10pt}
\begin{figure}
    \centering
    \includegraphics[width=0.63\linewidth]{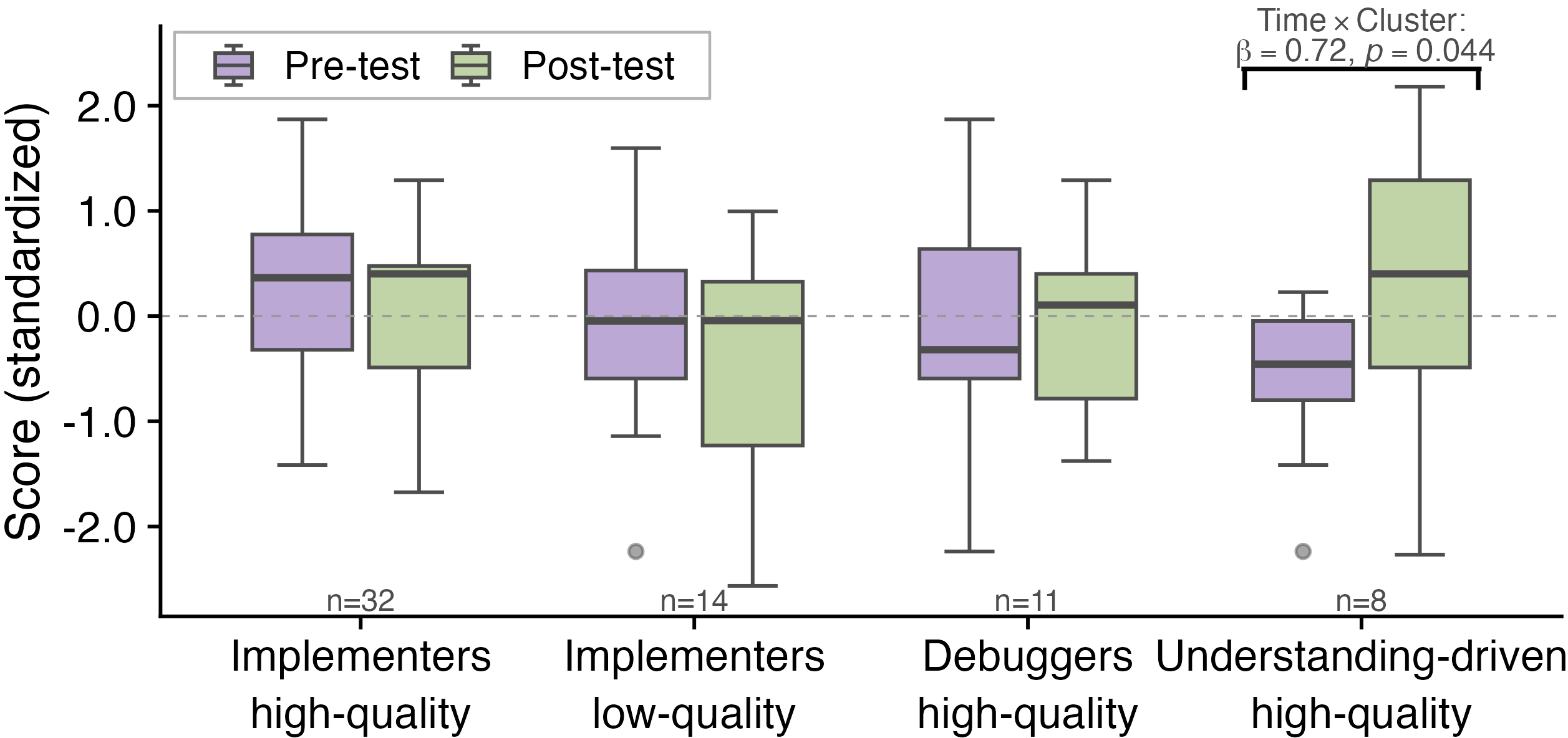}
        \caption{Learning outcomes in S3 by prompting-pattern cluster. 
Only the significant cluster is indicated, showing a significant interaction with time in repeated-measures mixed-effects model and greater learning relative to the grand mean across clusters (sum contrasts).}
    \label{fig:prompting_s3}
\end{figure}

\vspace{-9pt}
\paragraph{Task performance:} A Kruskal--Wallis test, complemented by regression analyses, revealed no significant differences in task performance between prompting-pattern clusters. Thus, there were comparable task-level outcomes across different ways of interacting with the LLM.

\vspace{-9pt}
\paragraph{Learning outcomes:} In contrast, learning outcomes differed across prompting patterns.
A repeated-measures mixed-effects regression predicting test scores from time (pre-post), prompting-pattern cluster, and their interaction revealed a significant interaction for the \textit{Understanding-driven prompters with high-quality prompts} cluster (Table~\ref{tab:s3_learning_regression}). Students in this cluster showed significantly greater learning gains than the grand mean across clusters ($\beta = 0.72$, $p = 0.044$), see Figure   \ref{fig:prompting_s3}. No other cluster showed a significant interaction with time, indicating that learning outcomes were associated with understanding-oriented prompting rather than with prompt quality or implementation-focused prompting alone. These results align with findings from prior studies \cite{brender2025structured}.

\begin{table}[htbp]
\vspace{-10pt}
\caption{Mixed effects repeated-measures regression of test scores by Time (pre--post) and Clusters, with students as a random effect. The individual clusters' effects are compared to the grand mean of all clusters using the sum of contrasts. Reference levels: Time = pre-test; Estimates are standardized coefficients ($\beta$).}
\label{tab:s3_learning_regression}
\centering
\begin{tabular}{lccc}
\toprule
Predictors & Estimates & CI (95\%) & $p$ \\
\midrule
(Intercept) & -0.16 & -0.43 -- 0.10 & 0.235 \\
Time [post-test] & 0.12 & -0.22 -- 0.47 & 0.484 \\ 
Implementers with low-quality & 0.03 & -0.42 -- 0.47 & 0.907 \\ 
Debuggers with high-quality & 0.02 & -0.47 -- 0.50 & 0.952 \\ 
Understanding-driven prompters with high-quality & -0.47 & -1.01 -- 0.07 & 0.089 \\ 
Time [post-test] : Implement. with low-quality & -0.48 & -1.05 -- 0.10 & 0.105 \\
Time [post-test] : Debuggers with high-quality  & -0.03 & -0.66 -- 0.59 & 0.915 \\
Time [post-test] : Understand.-driven with high-quality & {0.72}$^*$ & 0.02 -- 1.42 & {0.044} \\
\midrule
\multicolumn{4}{l}{Observations (students): 130 (65) \hspace{0.0cm} $R^2_\mathrm{marginal}$ / $R^2_\mathrm{conditional}$: 0.083 / 0.222 \quad $^* p \le 0.05$}\\ %\hspace{0.3cm}  $^* p \le 0.05$\\
\midrule
%\multicolumn{4}{c}{\quad $^* p \le 0.05$ \quad $^{**} p \le 0.01$ \quad $^{***} p \le 0.001$} \\
\end{tabular}
\vspace{-20pt}
\end{table}

\vspace{-7pt}\subsection{Post-intervention effects}\vspace{-3pt}

After finding that students using the SG-tutor achieved greater learning gains during the intervention while showing comparable task performance, we next examined whether these differences extended to students’ use of an unconstrained LLM in the course project.

\vspace{-12pt}
\subsubsection{Prompting patterns in the project.} 
Using prompt data from the course project, during which all students interacted with an unconstrained version of the LLM, we examined whether prompting patterns differed by intervention condition.
A cluster analysis of project prompts (Section~\ref{clustering}) identified three prompting patterns, labeled based on the high-low distribution of prompt types and quality:
(i) \textit{Implementers} with high-quality prompts ($n=25$),
(ii) \textit{Understanding-driven prompters} with high-quality prompts ($n=17$), and
(iii) \textit{Debuggers} with low-quality prompts ($n=10$).

Unlike in session S3, the distribution of students across clusters differed significantly by condition ($\chi^2=8.14$, $p=0.017$).
While the number of \textit{Implementers} was similar across conditions ($n_{SG}=14$, $n_{PR}=11$), students from the SG condition were more likely to enact \textit{Understanding-driven} prompting patterns ($n_{SG}=13$, $n_{PR}=4$), whereas students from the PR condition were more likely to enact \textit{Debugger} patterns ($n_{PR}=8$, $n_{SG}=2$).

These results suggest that participation in the SG condition was associated with the adoption of prompting practices oriented toward conceptual understanding that transferred to an unconstrained LLM-use context.

\vspace{-12pt}
\subsubsection{Prompting patterns and project understanding.} Models examining the link between condition and project understanding score showed that groups with a higher number of students who used the unconstrained course chatbot in the project tended to achieve higher understanding grades, regardless of condition. 

However, analyses of the relation of prompting patterns with group-level project understanding revealed significant differences. Specifically, application of the regression model on group-level project understanding scores, as explained in section \ref{sec:data_analysis}, showed that neither the total number of intervention students in a group nor the number of intervention students without chatbot use was significantly related to group understanding.

In contrast, prompting patterns were predictive of project understanding. Groups with a higher number of \textit{Understanding-driven prompters} achieved significantly higher understanding grades than groups with more \textit{Debuggers} ($\beta=1.30$, $p=0.025$), whereas the difference between \textit{Implementers} and \textit{Debuggers} was not significant ($\beta=0.76$, $p=0.19$).
Overall, the model explained 33\% of the variance in group-level understanding (adjusted $R^2=0.22$).

This analysis suggests that the association between the SG condition and higher project understanding operates through the prompting practices students enact during unconstrained LLM use, rather than through participation in the intervention alone.

\vspace{-7pt}\subsection{Students' perception of the different tutors}\vspace{-3pt}

Despite the positive effects associated with the SG-tutor, students in the PR condition reported more favorable perceptions of their tutor than students in the SG condition, with respect to perceived effectiveness for solving the task (Kruskal-Wallis $H=482$, $p_{\text{corrected}}=0.09$, $d=0.5$), learning ($H=449$, $p_{\text{corr}}=0.045$, $d=0.5$), and writing useful prompts ($H=440$, $p_{\text{corr}}=0.045$, $d=0.6$).

This difference may reflect more passive interaction in the PR condition: in session S3, 17 of 35 tutor-suggested prompt refinements were directly copy--pasted rather than revised.

These findings point to a trade-off between perceived usability and learning effectiveness, consistent with prior work showing that directive support can feel helpful while fostering shallower engagement \cite{brender2025structured}.

\vspace{-7pt}\section{Discussion and Conclusion}\vspace{-3pt}

This study examined how different forms of AI-tutor support shape students’ engagement with large language models (LLMs), and how these engagement patterns relate to task performance and learning during guided activities and beyond. We compared a Socratic-Guidance tutor that prompts reflective questioning with a Prompt-Refinement tutor that targets prompt formulation. Across the three research questions, the results indicate that tutor effects emerged over time, rather than through immediate differences in behavior or performance.

\vspace{-10pt}
\subsection{AI-tutor-supported engagement during the intervention}

\subsubsection{Prompting behaviors, task performance, and learning during guided use (RQ1).}
Analyses of prompting patterns during the guided lab sessions revealed no differences between conditions in terms of prompt type or prompt quality; at the level of observable prompting behaviors, students in both conditions engaged with the LLM in comparable ways.
In contrast, analyses of task performance (based on practice lab scores) and learning outcomes (based on pre-post tests) revealed a temporally evolving pattern.
Students in the SG condition initially showed lower task performance but similar learning gains relative to students in the PR condition. By the final session, students in the SG condition performed comparably while achieving higher learning gains. Socratic questioning requires learners to articulate reasoning and reflect on conceptual relations \cite{chen_exploring_2025}, which may initially reduce efficiency but supports progressively more productive engagement with the SG tutor’s reflective interaction style.

Perception data further highlight this distinction.
Despite stronger learning outcomes, students in the SG condition rated their tutor less favorably than students in the PR condition. These ratings may primarily reflect perceived ease of use, as the PR tutor afforded interaction patterns that made prompt reuse straightforward (e.g., direct copy–pasting of suggested refinements), reducing user effort and encouraging more passive engagement. From a learning-sciences perspective, this misalignment aligns with concerns about metacognitive offloading and overdependence, whereby supports that reduce interactional effort may feel efficient while fostering shallower engagement \cite{bastani2024generative}.

\vspace{-10pt}
\subsubsection{Prompting strategies and learning outcomes (RQ2).} Prior work shows that how students engage with LLMs matters for learning beyond task completion, particularly when interaction emphasizes conceptual explanation and sense-making rather than immediate solution generation \cite{brender2024s, brender2025structured}. 
Replicating this pattern, our results show that prompting patterns were unrelated to task performance but systematically associated with learning gains: students classified as \textit{Understanding-driven prompters using high-quality prompts} achieved significantly greater learning gains than students with other prompting patterns.

At the same time, a key tension emerges across RQ1 and RQ2. Although learning-relevant prompting patterns were clearly associated with improved learning, neither tutor led to greater adoption of such patterns during guided use, suggesting that inducing learning-conducive prompting strategies may require stronger or differently calibrated forms of guidance than those explored here.

\vspace{-10pt}
\subsection{Transfer of the AI-tutor's impact post-intervention (RQ3)}
\vspace{-5pt}

RQ3 examined whether the AI tutors influenced students’ later interaction with an unconstrained LLM during the course project, indicating more durable engagement.
First, the learning-relevant prompting pattern identified in RQ2 remained associated with higher group-level understanding beyond guided use.

Second, prompting patterns during the project differed by prior tutor condition. Students who had interacted with the SG tutor were more likely to enact \emph{understanding-driven prompting with high-quality prompts}, whereas students from the PR condition were more likely to rely on \emph{debugger-type prompting with lower-quality prompts}. Notably, these differences were not observed during the guided sessions, but emerged only in the unconstrained setting.

These findings resolve the tension identified in RQ1 and RQ2. While neither tutor produced immediate differences in prompting behavior during guided use, the SG tutor was associated with how students later engaged with LLMs once guidance was removed, suggesting a role of Socratic guidance that lies less in shaping prompts in the moment than in influencing how students develop approaches to problem solving and LLM use over time.

\vspace{-10pt}\subsection{Limitations}\vspace{-5pt}
This study involved a relatively small, domain-specific sample of graduate students, which limits generalizability. Participants self-selected into the study; however, comparisons were conducted between intervention conditions. Although we accounted for chatbot use and partial trace observability, unobserved differences in motivation or learning orientation may still have influenced prompting behavior and outcomes. The observed effects may depend on how the Socratic and prompt-refinement tutors were calibrated; different prompt designs or levels of guidance may lead to different interaction patterns and learning outcomes. Future studies with larger and more diverse populations are needed to validate these findings and to support stronger causal inferences.

\vspace{-10pt}\subsection{Conclusion and future work}
\vspace{-5pt}
This study highlights the importance of designing AI tutors that support students in learning how to learn with unconstrained LLMs, further confirming that learning outcomes depend less on immediate task performance or prompt efficiency than on understanding-oriented engagement.

A key contribution of this work is to show that differences between tutor designs emerge over time, rather than during guided interaction itself. While immediate prompting behavior did not differ, prior exposure to Socratic guidance was associated with more understanding-oriented engagement and higher understanding once students worked with LLMs independently, suggesting a developmental influence on how students learn to engage with these systems.

From a design perspective, the contrast between the two approaches is instructive. The SG tutor intervened by shaping how the LLM responds through questioning and reflection, whereas the PR tutor focused on improving the form of students’ prompts, such as clarity or granularity. The findings suggest that response-level, pedagogically oriented guidance may be more likely to influence how students later write prompts and engage with LLMs in unconstrained contexts than approaches centered on prompt efficiency alone. At the same time, these conclusions remain tentative, and further work is needed to explore alternative ways of fostering durable, learning-relevant engagement with LLMs.
\vspace{-15pt}

\vspace{-15pt}
\bibliographystyle{splncs04}
\bibliography{0-bib}

\end{document}